%% file: main.tex
\documentclass[runningheads, table]{llncs}

\input{format/packages}
\input{format/macros}

\title{
	Gaussian-Based and Outside-the-Box \\Runtime Monitoring Join Forces
	\thanks{The authors would like to thank all partners within the Hi-Drive project for their cooperation and valuable contribution. This project has received funding from the European Union's Horizon 2020 research and innovation program under grant agreement No 101006664 and the MUNI Award in Science and Humanities (MUNI/I/1757/2021) of the Grant Agency of Masaryk University. The sole responsibility of this publication lies with the authors. Neither the European Commission nor CINEA – in its capacity of Granting Authority – can be made responsible for any use that may be made of the information this document contains.}
}

\author{
	Vahid Hashemi \inst{1}
	\and Jan K\v{r}et\'{i}nsk\'{y}\inst{2, 3}\,\orcidID{0000-0002-8122-2881}	
	\and Sabine Rieder\inst{3,1,2}\,\orcidID{0009-0006-6397-3100}
	\and Torsten Schön\inst{4}
	\and Jan Vorhoff\inst{1,4}
}	
\institute{
	Audi AG, Ingolstadt, Germany \and
	Masaryk University, Brno, Czech \and 	
	Technical University of Munich, Munich, Germany \and
	AImotion Bavaria, Technische Hochschule Ingolstadt, Ingolstadt, Germany
	}

\begin{document}
	\begin{acronym}
		
		\acro{ood}[OOD]{Out-Of-Distribution}
		\acro{id}[ID]{In-Distribution}
		\acro{oms}[OMS]{Out-Of-Model-Scope}	
		\acro{id}[ID]{In-Distribution}
		\acro{nn}[NN]{Neural Network} 
		\acro{icad}[ICAD]{Inductive Conformal Anomaly Detection}
		\acro{fgsm}[FGSM]{Fast Gradient Sign Method}	
		\acro{sse}[SSE]{Sum of Squared Euclidean Distances}	
		\acro{tpr}[TPR]{True Positive Rate}
		
	\end{acronym}

\maketitle

\begin{abstract}
	Since neural networks can make wrong predictions even with high confidence, monitoring their behavior at runtime is important, especially in safety-critical domains like autonomous driving.
	In this paper, we combine ideas from previous monitoring approaches based on observing the activation values of hidden neurons. 
	In particular, we combine the Gaussian-based approach, which observes whether the current value of each monitored neuron is similar to typical values observed during training,
	and the Outside-the-Box monitor, which creates clusters of the acceptable activation values, and, thus, considers the correlations of the neurons' values.
	Our experiments evaluate the achieved improvement.
\end{abstract} 

\input{intro.tex}
\input{related-work.tex}
\input{preliminaries}
\input{methodology}
\input{evaluation}
\input{conclusion}

\newpage
\bibliographystyle{splncs04}
\bibliography{literature}

\appendix
\input{appendix.tex}

\end{document}

%% file: format/packages.tex
\usepackage{amsmath,amssymb,amsfonts,xcolor,graphicx,wrapfig,paralist}
\usepackage{algorithmicx,algorithm}
\usepackage[noend]{algpseudocode}
\usepackage{mathtools}
\usepackage{booktabs}
\usepackage{adjustbox}
\usepackage{xspace}
\usepackage{enumitem}
\usepackage{standalone}
\usepackage{pgfplots}
\usepackage{array}
\usepackage{csvsimple}
\usepackage{multirow}
\usepackage[utf8]{inputenc}
\usepackage{textgreek}
\usepackage[pdftex]{hyperref}
\usepackage{xcolor}
\usepackage{siunitx-v2}
\usepackage{bm}

\definecolor[named]{ACMBlue}{cmyk}{1,0.1,0,0.1}
\definecolor[named]{ACMYellow}{cmyk}{0,0.16,1,0}
\definecolor[named]{ACMOrange}{cmyk}{0,0.42,1,0.01}
\definecolor[named]{ACMRed}{cmyk}{0,0.90,0.86,0}
\definecolor[named]{ACMLightBlue}{cmyk}{0.49,0.01,0,0}
\definecolor[named]{ACMGreen}{cmyk}{0.20,0,1,0.19}
\definecolor[named]{ACMPurple}{cmyk}{0.55,1,0,0.15}
\definecolor[named]{ACMDarkBlue}{cmyk}{1,0.58,0,0.21}

\hypersetup{
	colorlinks,
	linkcolor=ACMPurple,
	citecolor=ACMPurple,
	urlcolor=ACMDarkBlue,
	filecolor=ACMDarkBlue}

\usepackage{algorithm}
\usepackage{algpseudocode}
\usepackage[nolist]{acronym}
\usepackage{breakcites} % break citations to a new line

\DeclareFixedFont{\ttb}{T1}{txtt}{bx}{n}{8} % for bold
\DeclareFixedFont{\ttm}{T1}{txtt}{m}{n}{8}  % for normal
\DeclareFixedFont{\tti}{T1}{txtt}{it}{n}{8}

\usepackage{color}
\definecolor{deepblue}{rgb}{0,0,0.5}
\definecolor{deepred}{rgb}{0.6,0,0}
\definecolor{deepgreen}{rgb}{0,0.5,0}

\usepackage{listings}

\newcommand\pythonstyle{\lstset{
		language=Python,
		basicstyle=\ttm,
		morekeywords={self},              % Add keywords here
		keywordstyle=\ttb\color{deepblue},
		emph={MyClass,__init__},          % Custom highlighting
		emphstyle=\ttb\color{deepred},    % Custom highlighting style
		stringstyle=\color{deepgreen},
		commentstyle=\tti\color{purple!60!black},
		frame=tb,                         % Any extra options here
		showstringspaces=false
}}

\lstnewenvironment{python}[1][]
{
	\pythonstyle
	\lstset{#1}
}
{}

\newcommand\pythoninline[1]{{\pythonstyle\lstinline!#1!}}

\lstdefinestyle{customJ}{
  belowcaptionskip=1\baselineskip,
  breaklines=true,
  frame=L,
  xleftmargin=\parindent,
  language=Java,
  showstringspaces=false,
  basicstyle=\footnotesize\ttfamily,
  keywordstyle=\bfseries\color{green!40!black},
  commentstyle=\itshape\color{purple!40!black},
  identifierstyle=\color{blue},
  stringstyle=\color{orange},
}

\usepackage[capitalise]{cleveref}

\usepackage[compatibility=false]{caption}
\usepackage{subcaption}
\usepackage{placeins} % For FloatBarrier

\usepackage{booktabs, multirow} % for borders and merged ranges

\usepackage[T1]{fontenc}
\usepackage{pifont}

\newcolumntype{R}[2]{%
	>{\adjustbox{angle=#1,lap=\width-(#2)}\bgroup}%
	l%
	<{\egroup}%
}
\newcommand*\rot{\multicolumn{1}{R{30}{1em}}}% no optional argument here, please!

%% file: intro.tex
\section{Introduction}
\acresetall

\emph{\acp{nn}} show impressive results on a variety of computer vision tasks \cite{Chen-2018-Semantic-Segmentation,Liu-2019-Object-Detection,Tan-2020-NN-Classification,Zhao-2020-Depth-Estimation}.
However, studies demonstrate that even state-of-the-art \acp{nn} experience reduced accuracy when processing so-called \emph{\ac{ood} data}, which follows a different distribution than the \emph{\ac{id} data} used during training \cite{ovadia-2019-performance-drop-on-ood2,shafaei-2019-performance-drop-on-ood3,recht-2019-performance-drop-on-ood}.
Furthermore, even very well-trained \acp{nn} fail to reach perfect accuracy on the testing data, although it is \ac{id}.
Such unavoidable inaccuracy poses a significant challenge in safety-critical domains like autonomous driving and
highlights the need for runtime monitors that hint at uncertain, and, thus, possibly erroneous, predictions.

\emph{\ac{oms} detection}~\cite{Guerin-2023-OODNotAllYouNeed} is the task of identifying inputs leading to incorrect predictions.
Monitors targeting the \ac{oms} setting aim to identify \ac{ood} and \ac{id} data for which the \ac{nn} produces incorrect results, but, in contrast to \ac{ood} detection, should not notify of correctly processed \ac{ood} data.
Since the primary objective of runtime monitoring 
is identifying erroneous outcomes, 
\ac{oms} detection is more adequate in a safety-critical context 
than \ac{ood} detection.
Nevertheless, \ac{ood} detection may serve as a valuable proxy to attain the ultimate goal of \ac{oms} detection.

In this work, we combine ideas of two existing lightweight \ac{ood} detection approaches, namely the Gaussian monitor~\cite{gaussianMonitor} and the Outside-the-Box monitor~\cite{Henzinger-2020-AbstractionBasedMonitor} (short: Box monitor)
and evaluate the combination for \ac{oms} detection. 
Both approaches base their \ac{ood} decision on the activation values of an inner, typically the penultimate, layer.
The Box monitor records the activation values of known data and creates areas of previously seen activation value vectors in the form of unions of hyperrectangles.
Inputs producing values inside the box are assumed to be safe.
The Gaussian monitor \cite{gaussianMonitor} instead models the activation values of a neuron as a Gaussian distribution and computes intervals that contain \emph{likely} values of this distribution.
Consequently, it ignores outliers in training and warns against rare values.
However, the Gaussian monitor does not use the information of correlations between neurons used in the Box monitor as it,
technically, only considers one hyperrectangle.
Our combination of the two approaches enriches the Gaussian monitor with information about these correlations among
the individual neurons.
As the amount of computations increases with the number of monitored neurons,
we decrease the number of monitored neurons by using partial gradient descent.

Our basic experiments in the \ac{oms} setting show that considering the correlations of neurons improves the detection capabilities of the vanilla Gaussian monitor for the complex \cifarten \cite{cifar10} dataset.
In contrast, on the simpler \gtsrb \cite{gtsrb} dataset, the combination of the Box monitor and the Gaussian monitor achieves similar results as the latter.
Interestingly, reducing the number of monitored neurons does not drastically decrease performance for the combined monitor, which altogether allows for more efficient runtime monitoring.

\paragraph{Our contribution} can be summarized as follows:
\begin{itemize}
	\item We extend the monitor presented in \cite{gaussianMonitor} to benefit from clustering of activation values from the Box monitor \cite{Henzinger-2020-AbstractionBasedMonitor}.
	\item As the computation requirements increase with the size of the \ac{nn}, we investigate monitoring only the most relevant neurons of the \ac{nn}.
	\item We evaluate the presented monitoring approaches for \ac{oms} detection on the \cifarten and \gtsrb datasets.
\end{itemize}

%% file: related-work.tex
\section{Related Work}
\label{sec:rw}

As our work focuses on detecting \ac{oms} data based on the activation values of hidden layers in the \ac{nn},
we mention related work following the sane approach.
The survey by Yang et al. \cite{yang2021generalized} presents a more comprehensive overview of \ac{ood} detection approaches. 

Cheng et al. \cite{Cheng-2019-NeuronActivationPatternsMonitor} focus on the status (equal to zero or above zero) of \relu neurons on known safe data
and store it as a ``pattern''.
Patterns observed at runtime are accepted if they are close to already stored patterns. 
Henzinger et al.~\cite{Henzinger-2020-AbstractionBasedMonitor} consider the activation values 
and build safe areas in the form of hyperrectangles based on the observed values.
Lukina et al.~\cite{lukina2021into} extend the approach to incrementally adapt at runtime and create a quantitative measure of how unknown a new sample is. 
Hashemi et al. \cite{gaussianMonitor} model the activation values of a neuron with a Gaussian distribution and build an interval containing common activation values centered around the mean of the distribution. 
Sun et al. \cite{ood-nearest-neighbor} use the distance of an activation value to its k-nearest neighbor in a set of previously recorded activation values as \ac{ood} score. 
Morteza and Li \cite{morteza2022provable} assume
that the activation values of \ac{id} data follow a multivariate Gaussian distribution
and measure the sum of values proportional to each class log-likelihood of the in-distribution data. 
Lee et al. \cite{mahalanobis-monitor} fit a class-dependent multivariate Gaussian distribution to the observed activation values of a softmax classifier and compute a score based on the Mahalanobis-distance.
Corbi\`ere et al. \cite{Corbi-2019-QualityScores} suggest using an \ac{nn} trained on the neuron activation values of the penultimate layer as a monitor.
This \ac{nn} is supposed to predict the probability of the monitored \ac{nn} making a correct prediction for the input.

%% file: preliminaries.tex
\section{Preliminaries} \label{sec:prel}

\subsection{Neural Networks}
An \ac{nn} consists of $m$ layers $L_1$ to $L_m$
and each layer $l$ contains $s_l$ 
neurons.
Each neuron of layer $l>1$ first computes its pre-activation value $\mathbf{z}$ as weighted sum of activation values from neurons of the precious layer.
The neuron's activation value $\mathbf{a}$ is obtained by applying an 
activation function $f_l:\mathbb{R} \rightarrow \mathbb{R}$ to the pre-activation value.
With a slight abuse of notation, we write $f_l(\mathbf{z})$
when we refer to component-wise application of $f_l$ on $\mathbf{z}$.
We formalize the computations 
for a layer $l$ with weight matrix $W_l \in \mathbb{R}^{s_{l-1}\times s_l}$, bias $\mathbf{b_l}$ and input $\mathbf{x} \in D \subseteq \mathbb{R}^n$ as:
\begin{align*}
	\mathbf{a}_0 &= \mathbf{x} &
	\mathbf{z}_l &= \mathit{W}_{l}^{T}\,\mathbf{a}_{l-1} + \mathbf{b_l} & 
	\mathbf{a}_l &= f_l(\mathbf{z}_l) &
\end{align*}
The last layer provides the classification result, meaning the mapping of datapoint $\mathbf{x}$ 
to a set of $r$ different labels $Y = \{y_1, \ldots, y_r 
\}$.

\subsection{Monitors}
\inlineheadingbf{Gaussian Monitor \cite{gaussianMonitor}}
The approach consists of three steps: Training of the monitor, threshold setting and runtime evaluation.
In the training process, for each neuron and class, a Gaussian distribution is fitted to the neuron's activation values observed for several \ac{id} inputs.
We then compute an interval for each neuron containing approximately $95\%$ of recorded activation values based on the empirical rule~\cite{Wackerly-2014-Statistics}.
Due to the use of the empirical rule, it is barely ever the case that all neurons produce values contained in their respective intervals.
This is rectified by setting a threshold on the number of neurons that need to produce such values. 
A higher threshold will lead to the monitor classifying more images as \ac{ood} and vice versa.
We consider a set of known safe images not used for training the monitor for setting this threshold.
Lastly, when the \ac{nn} receives a new sample as input, the neurons' activation values and the predicted class are observed.
Each neuron votes if its value lies inside the bounds for the predicted class, and the outcome of the vote is compared to the threshold.

\inlineheadingbf{Box Monitor \cite{Henzinger-2020-AbstractionBasedMonitor}}
Henzinger et al. suggest using multidimensional boxes for monitoring, one set of boxes for each class.
The boxes are computed based on activation values observed on \ac{id} inputs.
These values are clustered with the K-Means algorithm and each cluster results in one box. 
As there exists a trade-off between the number of correctly identified \ac{ood} points and wrongly identified \ac{id} data, it is possible to modify the coarseness of the boxes by enlarging them by a factor of $\gamma$. 
During runtime, the authors expect \ac{id} data to produce activation vectors contained in the boxes and \ac{ood} data to produce values outside of it.

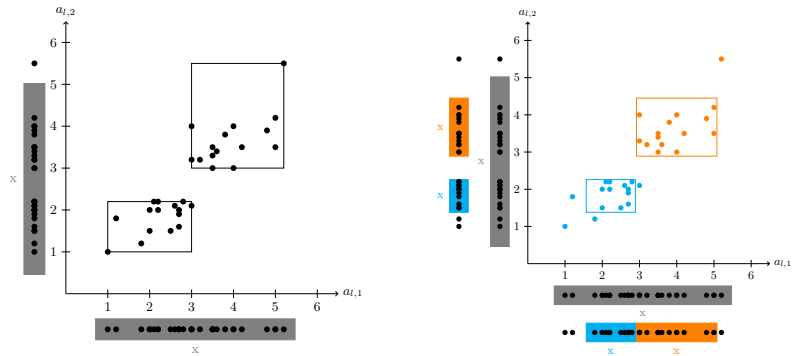
\begin{figure}
	\centering
	\begin{subfigure}{0.4\textwidth}
		\centering
		\input{figures/box-vs-gauss_v2}
		\caption{Comparison of the  Gaussian monitor and the Box monitor. The figure is based on \cite{gaussianMonitor}.\\}
		\label{fig:box_vs_gaussian}
	\end{subfigure}%
\hfil
	\centering
	\begin{subfigure}{0.4\textwidth}
		\centering
		\input{figures/gaussian_vs_clustered_gaussian}
		\caption{Comparison of Gaussian monitor and the clustered Gaussian monitor. The colors denote the assigned point cluster. }
		\label{fig:gaussian_vs_clustered_gaussian}
	\end{subfigure}
	\caption{Comparison of different monitors. The axes represent the neurons and their activation values, and the points are the vectors of activation values observed for a particular input. Boxes depict safe areas of the Box and the combined monitor. The points below each axis show the activation values mapped to the specific axis with $x$ highlighting the mean. 
		The boxes below the axes depict the activation values the Gaussian monitor accepts.}
\end{figure}

\inlineheadingbf{Difference between Gaussian and Box Monitor} 
\Cref{fig:box_vs_gaussian} depicts the differences between the two described monitors.
While the Box Monitor considers the relationship between neurons, the Gaussian monitor models the behavior of each single neuron more precisely and is less affected by outliers.

\subsection{Neuron Selection via Gradient Analysis}

Due to the additional complexity each new neuron brings to monitoring, Cheng et al. \cite{Cheng-2019-NeuronActivationPatternsMonitor} propose, among other things, only 
observing the most relevant neurons in the monitored layer $L_l$.
They identify these neurons through the absolute value of the partial gradient 
$\abs{\frac{\partial a_{m,j}}{\partial a_{l,t}}}$
for activation value $a_{m,j}$ of neuron $N_{m,j}$ in the output layer $L_m$, and the activation $a_{l,t}$ of a neuron $N_{l,t} \in L_l$. 
The partial derivatives depend on the activation function's derivatives and vary for each input. 
A special case is monitoring last hidden layer without any non-linear computation.
In this particular case, as pointed out by Cheng et al. \cite{Cheng-2019-NeuronActivationPatternsMonitor}, the partial derivatives are the weights between the two layers.
The paper is unclear on handling the case with different partial derivatives for various inputs.

%% file: figures/box-vs-gauss_v2.tex
\resizebox{\textwidth}{!}{
\begin{tikzpicture}
% Define the stretch factor
  \pgfmathsetmacro{\stretch}{0}
% Define a list of values

\def\xvalues{2, 2.5, 2.2, 2.8, 1, 2, 1.2, 1.8, 4, 3.5, 4.2, 4, 3, 3.8, 3.2, 3.5, 5, 4.8, 5.2, 5, 3, 3.5, 3.6, 3, 2.7, 2.7, 2.2, 2.6, 2.7, 2.1}
\pgfmathsetmacro{\xmean}{3.09}
\pgfmathsetmacro{\xstddev}{1.19}

\def\yvalues{2, 1.5, 2, 2.2, 1, 1.5, 1.8, 1.2, 4, 3, 3.5, 3, 4, 3.8, 3.2, 3.5, 4.2, 3.9, 5.5, 3.5, 2.1, 3.3, 3.4, 3.2, 1.9, 1.6, 2.2, 2.1, 2, 2.2}
\pgfmathsetmacro{\ymean}{2.74}
\pgfmathsetmacro{\ystddev}{1.14}

    %gaussian y
   % Draw a red box to the left of the axes
   \pgfmathsetmacro\ydown{\ymean - 2 * \ystddev} % Calculate the y-coordinate
   \pgfmathsetmacro\yup{\ymean + 2 * \ystddev} % Calculate the y-coordinate
   
  \draw[gray, fill=gray](-1,\ydown) rectangle (-0.5,\yup);

    %gaussian x 
   % Draw a red box to the left of the axes
   \pgfmathsetmacro\xdown{\xmean - 2 * \xstddev} % Calculate the y-coordinate
   \pgfmathsetmacro\xup{\xmean + 2 * \xstddev} % Calculate the y-coordinate
  \draw[gray, fill=gray] (\xdown,-0.6) rectangle (\xup,-1.1);

  % Draw axes
  \draw[->] (0,0) -- (6.5,0) node[right] {$a_{l,1}$};
  \draw[->] (0,0) -- (0,6.5) node[above] {$a_{l,2}$};

      % X-axis ticks and labels
  \foreach \x in {1,2,...,6}
  {
    \draw (\x,0.1) -- (\x,-0.1) node[below] {$\x$};
  }

  % Y-axis ticks and labels
  \foreach \y in {1,2,...,6}
  {
    \draw (0.1,\y) -- (-0.1,\y) node[left] {$\y$};
  }

   % Gaussian points x
  \foreach \x in \xvalues
  {
    \fill[black] (\x,-0.85) circle (2pt);
  }

   % Gaussian points y
  \foreach \y in \yvalues
  {
    \fill[black] (-0.75,\y) circle (2pt);
  }

  % x mean
  \node[gray] at (\xmean,-1.35) {x};

  % y mean
  \node[gray] at (-1.25,\ymean) {x};

\foreach \x [count=\c,evaluate=\c as \y using {{\yvalues}[\c-1]}]  in \xvalues 
  {
    \fill (\x,\y) circle (2pt);
  }

\draw[black] (3 - \stretch,3 - \stretch) rectangle (5.2 + \stretch,5.5 + \stretch);
\draw[black] (1 - \stretch,1 - \stretch) rectangle (3 + \stretch,2.2 + \stretch);

\end{tikzpicture}}

%% file: figures/gaussian_vs_clustered_gaussian.tex
\resizebox{\textwidth}{!}{
\begin{tikzpicture}
% Define the stretch factor
  \pgfmathsetmacro{\stretch}{0}
% Define a list of values

\def\xvalues{2, 2.5, 2.2, 2.8, 1, 2, 1.2, 1.8, 4, 3.5, 4.2, 4, 3, 3.8, 3.2, 3.5, 5, 4.8, 5.2, 5, 3, 3.5, 3.6, 3, 2.7, 2.7, 2.2, 2.6, 2.7, 2.1}
\pgfmathsetmacro{\xmean}{3.09}
\pgfmathsetmacro{\xstddev}{1.19}

\def\xvaluesone{2, 2.5, 2.2, 2.8, 1, 2, 1.2, 1.8, 3, 2.7, 2.7, 2.2, 2.6, 2.7, 2.1}
\pgfmathsetmacro{\xmeanone}{2.23}
\pgfmathsetmacro{\xstddevone}{0.33}

\def\xvaluestwo{4, 3.5, 4.2, 4, 3, 3.8, 3.2, 3.5, 5, 4.8, 5.2, 5, 3, 3.5, 3.6}
\pgfmathsetmacro{\xmeantwo}{4}
\pgfmathsetmacro{\xstddevtwo}{0.54}

\def\yvalues{2, 1.5, 2, 2.2, 1, 1.5, 1.8, 1.2, 4, 3, 3.5, 3, 4, 3.8, 3.2, 3.5, 4.2, 3.9, 5.5, 3.5, 2.1, 3.3, 3.4, 3.2, 1.9, 1.6, 2.2, 2.1, 2, 2.2}
\pgfmathsetmacro{\ymean}{2.74}
\pgfmathsetmacro{\ystddev}{1.14}

\def\yvaluesone{2, 1.5, 2, 2.2, 1, 1.5, 1.8, 1.2, 2.1, 1.9, 1.6, 2.2, 2.1, 2, 2.2}
\pgfmathsetmacro{\ymeanone}{1.82}
\pgfmathsetmacro{\ystddevone}{0.22}

\def\yvaluestwo{4, 3, 3.5, 3, 4, 3.8, 3.2, 3.5, 4.2, 3.9, 5.5, 3.5, 3.3, 3.4, 3.2}
\pgfmathsetmacro{\ymeantwo}{3.67}
\pgfmathsetmacro{\ystddevtwo}{0.39}

    %gaussian y
   % Draw a red box to the left of the axes
   \pgfmathsetmacro\ydown{\ymean - 2 * \ystddev} % Calculate the y-coordinate
   \pgfmathsetmacro\yup{\ymean + 2 * \ystddev} % Calculate the y-coordinate
   
  \draw[gray, fill=gray](-1,\ydown) rectangle (-0.5,\yup);

    %gaussian x 
   % Draw a red box to the left of the axes
   \pgfmathsetmacro\xdown{\xmean - 2 * \xstddev} % Calculate the y-coordinate
   \pgfmathsetmacro\xup{\xmean + 2 * \xstddev} % Calculate the y-coordinate
  \draw[gray, fill=gray] (\xdown,-0.6) rectangle (\xup,-1.1);
	  
    %gaussian x  cluster 2
   % Draw a red box to the left of the axes
      \pgfmathsetmacro\xdown{\xmeantwo - 2 * \xstddevtwo} % Calculate the y-coordinate
   \pgfmathsetmacro\xup{\xmeantwo + 2 * \xstddevtwo} % Calculate the y-coordinate
  	\draw[orange, fill=orange] (\xdown,-1.6) rectangle (\xup,-2.1);
  	\pgfmathsetmacro\ydown{\ymeantwo - 2 * \ystddevtwo} % Calculate the y-coordinate
  	\pgfmathsetmacro\yup{\ymeantwo + 2 * \ystddevtwo} % Calculate the y-coordinate
  	
  	\draw[orange, fill=orange](-2.1,\ydown) rectangle (-1.6,\yup);
  	
	\draw[orange, fill=none] (\xdown,\ydown) rectangle (\xup,\yup);
	  
    %gaussian x cluster 1
   % Draw a red box to the left of the axes
    \pgfmathsetmacro\xdown{\xmeanone - 2 * \xstddevone} % Calculate the y-coordinate
   	\pgfmathsetmacro\xup{\xmeanone + 2 * \xstddevone} % Calculate the y-coordinate
   	
   	\pgfmathsetmacro\ydown{\ymeanone - 2 * \ystddevone} % Calculate the y-coordinate
   	\pgfmathsetmacro\yup{\ymeanone + 2 * \ystddevone} % Calculate the y-coordinate
   	
   	\draw[cyan, fill=cyan](-2.1,\ydown) rectangle (-1.6,\yup);
   
  	\draw[cyan, fill=cyan] (\xdown,-1.6) rectangle (\xup,-2.1);
  
  	\draw[cyan, fill=none] (\xdown,\ydown) rectangle (\xup,\yup);
  
  % Draw axes
  \draw[->] (0,0) -- (6.5,0) node[right] {$a_{l,1}$};
  \draw[->] (0,0) -- (0,6.5) node[above] {$a_{l,2}$};

      % X-axis ticks and labels
  \foreach \x in {1,2,...,6}
  {
    \draw (\x,0.1) -- (\x,-0.1) node[below] {$\x$};
  }

  % Y-axis ticks and labels
  \foreach \y in {1,2,...,6}
  {
    \draw (0.1,\y) -- (-0.1,\y) node[left] {$\y$};
  }

   % Gaussian points x
  \foreach \x in \xvalues
  {
    \fill[black] (\x,-0.85) circle (2pt);
  }
     % Gaussian points x cluster 2
  \foreach \x in \xvaluestwo
  {
    \fill[black] (\x,-1.85) circle (2pt);
  }
     % Gaussian points x cluster 1
  \foreach \x in \xvaluesone
  {
    \fill[black] (\x,-1.85) circle (2pt);
  }

   % Gaussian points y
  \foreach \y in \yvalues
  {
    \fill[black] (-0.75,\y) circle (2pt);
  }

 % Gaussian points y cluster 2
  \foreach \y in \yvaluestwo
  {
    \fill[black] (-1.85,\y) circle (2pt);
  }

   % Gaussian points y cluster 1
  \foreach \y in \yvaluesone
  {
    \fill[black] (-1.85,\y) circle (2pt);
  }

  % x mean
  \node[gray] at (\xmean,-1.35) {x};
  % x mean cluster 2
  \node[orange] at (\xmeantwo,-2.35) {x};
  % x mean cluster 1
  \node[cyan] at (\xmeanone,-2.35) {x};

  % y mean
  \node[gray] at (-1.25,\ymean) {x};
  % y mean cluster 2
  \node[orange] at (-2.35,\ymeantwo) {x};
  % y mean cluster 1
  \node[cyan] at (-2.35,\ymeanone) {x};

% two dimensional points cluster 1
\foreach \x [count=\c,evaluate=\c as \y using {{\yvaluesone}[\c-1]}]  in \xvaluesone
  {
    \fill[cyan] (\x,\y) circle (2pt);
  }

 % cluster 2
 \foreach \x [count=\c,evaluate=\c as \y using {{\yvaluestwo}[\c-1]}]  in \xvaluestwo
   {
    \fill[orange] (\x,\y) circle (2pt);
  }

\end{tikzpicture}}

%% file: methodology.tex
\section{Methodology}
\label{sec:method}

\subsection{Combining Gaussian Monitor and Box Monitor}
We combine the Box monitor and the Gaussian monitor to obtain a monitor robust to outliers (like the Gaussian monitor) and aware of neuron correlations (like the Box monitor).
The monitor is trained based on an observation of $k$ activation vectors for each class $c$.
These $k$ activation vectors are obtained from correctly classified \ac{id} data and are clustered using K-Means similar to the creation of the Box monitor.
Overall, we obtain several clusters per class. 
The Gaussian Monitor is called on each cluster separately and computes a Gaussian distribution for each neuron and its activation values belonging to the clusters.
Based on the distribution, intervals containing $95\%$ of the data are computed as before and boxes are build based on these intervals.
\cref{fig:gaussian_vs_clustered_gaussian} shows the boxes for different clusters (depicted by different colors).
For an input to be accepted, its activation value needs to be contained in one of the boxes.
While the Gaussian monitor from \cref{fig:box_vs_gaussian} would have accepted an activation value of $(2,4)$, the new monitor will mark it correctly as previously unseen. 
Furthermore, the combined monitor is more sensitive to cluster outliers than the Box monitor.
The sensitivity can be fine tuned by setting a threshold on how many neurons need to produce values contained inside the intervals of a specific cluster.
Due to the new structure of the monitor, we use a procedure similar to Henzinger et al. \cite{Henzinger-2020-AbstractionBasedMonitor} for detecting outliers at runtime.
When a new input is observed, we compute the Euclidean distance of its activation value vector to the centroids of each cluster and compare to the closest ones.
For each cluster, each neuron still votes on whether its activation is in the interval and thus valid and an alarm will be triggered if the number is lower than the threshold.

One can also use a multivariate Gaussian distribution for each cluster and compute the Mahalanobis distance similar to the suggestion by Lee et al. \cite{mahalanobis-monitor}.

\subsection{Neuron Selection for Monitoring}\label{subsubsec:neuron_selection_monitored_nn}
To improve the runtime performance of our proposed monitors, we adhere to the suggestion made by Cheng et al. \cite{Cheng-2019-NeuronActivationPatternsMonitor} to only observe the neurons with the most significant impact on the expected class score determined by partial gradient analysis.
We compute a score $abs\_score^l_{j,t}$ for a particular class $y_j$ and a neuron $N_{l,t}$ of layer $l$ based on the partial derivative of several known safe inputs.
We define $S_j\subset D$ as a set of known safe inputs of class $j$.
To calculate the $abs\_score^l_{j,t}$ for the $N_{l,t}$, we sum up the absolute value of the partial gradient $\frac{\partial a_{m,j}(x_i)}{\partial a_{l,t}(x_i)}$ for each input vector $\mathbf{x}_i \in S_j$, where $a_{m,j}(x_i)$ represents the activation value of neuron $N_{m,j}$ when the \ac{nn} is applied to $x_i$:
$$
abs\_score^l_{j,t} = \sum_{i=1}^{|S_j|} \left|\frac{\partial a_{m,j}(x_i)}{\partial a_{l,t}(x_i)}\right|.
$$
While we can apply this technique of selection relevant neurons to the \ac{nn} we want to monitor, there is also the possibility to use a \emph{monitoring \ac{nn}}~\cite{Corbi-2019-QualityScores}.
In this case, we first train an \ac{nn} on the (pre-)activation values of the original \ac{nn} for each class to detect potentially misclassified images. 
Given the set of safe activation values $A^l_j$ for class $j$ and layer $l$ and the set of activation $\bar{A}^l_j$ for images of the training data that do not lead to the class prediction, an \ac{nn}, defined as $NET^l_j$, is created to output a score between $0$ and $1$ that encodes the confidence the monitoring \ac{nn} has that the prediction made by the monitored \ac{nn} is correct.
Basing our neuron selection on the partial gradient of such a monitoring \ac{nn} uses the fact that this \ac{nn} has obtained 
a deeper understanding of the original \ac{nn}.

%% file: evaluation.tex
\section{Evaluation}
\label{sec:ev}

In our experiments, we evaluate (RQ1) the influence of neuron selection, (RQ2) the combination of the multivariate Gaussian monitor with clustering and (RQ3) the overall performance of our suggested combination of methods.

\subsection{Experimental Setup and Implementation}

\inlineheadingbf{Datasets}
For our experiments, we use two datasets, GTSRB \cite{gtsrb}, a dataset consisting of German traffic signs, and \cifarten \cite{cifar10}, which contains objects from different classes like plane, car, or bird.
We evaluate our proposed monitors 
with the Monitizer framework \cite{monitizer}, which allows us to implement a monitor and evaluate it on various types of \ac{ood} data.
As we are interested in \ac{oms} detection, we generalize this framework 
by excluding \ac{ood} images from evaluation if the monitored \ac{nn} still made a correct prediction and
adding a new category of misclassified \ac{id} images.
The \ac{oms} data classes contain images from an entirely new dataset, different noises and perturbations like light and contrast changes applied to known data, and objects not contained in the training data, but placed in a similar environment as the one known to the \ac{nn}.
\cref{app:datasets} contains a visualization of the datasets and \ac{oms} data.

\inlineheadingbf{Network Architectures}
For \cifarten images, we use an \ac{nn} consisting of 6 convolutional layers with \relu activation function and max-pooling operation followed by 4 linear layers with \relu activation function~\cite{Vishal-Ramesh-NN-Architecture-Cifar10}.
The \ac{nn} for \gtsrb consists of 2 convolutional layers with \relu activation and max-pooling operation followed by 3 linear layers with \relu activation function.
Both architectures are illustrated in \cref{app:networks}.

\inlineheadingbf{Preparation of Monitors}
For training the monitors, meaning generating the intervals and boxes, and threshold setting, we use the pre-activations of the \ac{nn} for correctly classified input images, following the suggestion of previous works~\cite{Cheng-2019-NeuronActivationPatternsMonitor,Henzinger-2020-AbstractionBasedMonitor}.  
We focus on pre-activation values as we have obtained better results on them compared to the activation values (see \cref{app:pre-act}).
The images for threshold setting have not been used for clustering and computing intervals. 
We adjust our monitors' thresholds to receive no alarm on $90\%$ of the correctly classified \ac{id} dataset. 

\inlineheadingbf{Quality Measure and Plots}
The plots show the \ac{tpr} (sometimes also called \emph{recall}) of our experiments, meaning the number of correct alarms divided by the number of \ac{oms} images for each of our \ac{oms} datasets.
This is the most relevant metric as our datasets only consist of \ac{oms} examples.
The monitors' performance on data within the model-scope is fixed by the threshold.

\subsection{Results}
\inlineheadingbf{RQ1: Neuron Selection}
\Cref{tab:neruon-selection} displays the \ac{tpr} for the clustered Gaussian monitor when monitoring the pre-activation values of all neurons in the last hidden layer versus monitoring only $75\%,\ 50\%$ or $25\%$ of the neurons for \cifarten. 
The results indicate that monitoring less neurons tends to decrease the \ac{tpr}. 
However, when comparing the selection of $100\%$ with the selection of only $25\%$, the decrease is mostly not significant, while 
reducing the number of monitored neurons by $75\%$ drastically reduces computation efforts.
\Cref{tab:neruon-selection} also displays the \ac{tpr} when selecting $75\%$ of the neurons for monitoring based on partial gradient descent of the monitoring \ac{nn}. 
Our monitoring \ac{nn} consists of 3 linear layers with \relu activation function followed by one layer with Sigmoid activation function.
It shows that the results obtained when using the gradients computed on the monitoring \ac{nn} perform slightly better in some settings than the ones for the original \ac{nn} but are mostly similar.
\input{plots/cifar10/neuron-selection.tex}

\inlineheadingbf{RQ2: Clustering the Multivariate Gaussian~\cite{mahalanobis-monitor}}
\input{plots/cifar10/multi-gauss-cluster.tex}
\Cref{tab:multi-gauss-cluster} displays the \ac{tpr} of the monitor when using a different number of clusters for the pre-activation values or the last hidden layer of the \cifarten \ac{nn}.
Using clusters or increasing their number reduces performance on all test settings. 

\inlineheadingbf{RQ3: Overall Results}
\input{plots/overall.tex}
\Cref{tab:overall} shows the \ac{tpr} for the different monitors using the best configurations  
for the \cifarten and the \gtsrb dataset.
We compare our monitors to the Gaussian monitor as Hashemi et al.~\cite{gaussianMonitor} demonstrated that it performs at least as well as the Box monitor.
Our proposed runtime monitors demonstrate better results than the Gaussian monitor in all our test settings by at least $10\%$ on the \cifarten dataset. 
Also on the \gtsrb dataset, the clustered Gaussian monitor performs at least as well as the Gaussian monitor~\cite{gaussianMonitor}, but the improvement is not as noticeable anymore.
All monitors attain notably higher \acp{tpr} when assessed on the \gtsrb dataset than on the \cifarten dataset.

\subsection{Discussion}
Combining the research findings of Hashemi et al. \cite{gaussianMonitor} and Henzinger et al. \cite{Henzinger-2020-AbstractionBasedMonitor} has demonstrated that pre-clustering the \ac{nn}'s activations before constructing the Gaussian monitor is superior to a Gaussian monitor without clusters. 
This is particularly evident when using slightly more complex datasets,
%and architectures, 
such as \cifarten.
On the less complex \gtsrb dataset the new monitor achieves at least similar results to the baseline method making our approach applicable to simpler datasets and beneficial for more complex datasets.
Clustering activation values for the multivariate Gaussian monitor~\cite{mahalanobis-monitor} did not improve performance.

By selecting and monitoring only the most relevant neurons, identified through partial gradient descent on several inputs, the monitor's performance is not drastically decreased.
This enables more efficient monitoring of activation values.

%% file: plots/cifar10/neuron-selection.tex
\begin{table}
	\caption{Comparison of the \ac{tpr} for the clustered Gaussian monitor with 3 Clusters when only using the $x\%$ most relevant neurons (partial gradient descent on the original or the monitoring \ac{nn}) for the pre-activation values of the last hidden layer of the \cifarten \ac{nn}.}
	\label{tab:neruon-selection}
	\centering
	\resizebox{0.9\textwidth}{!}{
		\begin{tabular}{l|r|r|r|r|r|r|r|r|r|r|r}
			Monitors & \rot{Wrong ID} & \rot{GTSRB} & \rot{DTD} & \rot{Gaussian} & \rot{SaltAndPepper} & \rot{Contrast} & \rot{GaussianBlur} & \rot{Invert} & \rot{Rotate} & \rot{Light} & \rot{Cifar100} \\
			\toprule
			Original NN &&&&&&&&&&&\\
			\quad 100\% & 21.71 & 39.00 & 15.00 & 19.49 & 26.20 & 32.42 & 17.02 & 21.76 & 20.74 & 24.58 & 20.00 \\
			\quad 75\% & 17.74 & 44.00 & 13.00 & 13.62 & 12.44 & 28.82 & 9.21 & 14.14 & 13.32 & 22.05 & 17.00 \\
			\quad 50\% & 20.05 & 41.00 & 21.00 & 16.43 & 17.35 & 31.32 & 11.25 & 16.86 & 17.11 & 24.36 & 18.00 \\
			\quad 25\% & 19.59 & 36.00 & 21.00 & 18.34 & 22.89 & 29.58 & 12.05 & 18.73 & 20.27 & 21.99 & 12.00 \\
			Monitoring NN &&&&&&&&&&&\\
			\quad 75\% & 17.67 & 39.00 & 17.00 & 14.75 & 18.51 & 28.42 & 12.62 & 15.81 & 15.07 & 20.77 & 14.00 \\
		\end{tabular}
	}
\end{table}

%% file: plots/cifar10/multi-gauss-cluster.tex
\begin{table}
	\caption{\ac{tpr} of the multivariate Gaussian monitor applied to the pre-activation values of the last hidden layer (\cifarten) with different numbers of clusters.}
	\label{tab:multi-gauss-cluster}
	\centering
	\resizebox{0.9\textwidth}{!}{
		\begin{tabular}{l|r|r|r|r|r|r|r|r|r|r|r}
			Monitors & \rot{Wrong ID} & \rot{GTSRB} & \rot{DTD} & \rot{Gaussian} & \rot{SaltAndPepper} & \rot{Contrast} & \rot{GaussianBlur} & \rot{Invert} & \rot{Rotate} & \rot{Light} & \rot{Cifar100} \\
			\toprule
			No \ Clusters & 24.21 & 57.00 & 12.00 & 18.93 & 15.64 & 42.48 & 12.47 & 21.99 & 20.57 & 34.47 & 25.00 \\
			2 \ Clusters & 20.05 & 51.00 & 10.00 & 15.84 & 13.03 & 37.54 & 9.62 & 17.96 & 17.21 & 29.93 & 24.00 \\
			3 \ Clusters & 16.20 & 43.00 & 8.00 & 12.90 & 10.63 & 33.48 & 7.34 & 15.03 & 14.29 & 25.19 & 21.00 \\
		\end{tabular}
	}
\end{table}

%% file: plots/overall.tex
\begin{table}
	\caption{Comparison of the different monitors on the \cifarten (top table) and the \gtsrb dataset (lower table). In the first 3 rows, we consider the pre-activation values of the last hidden layer. The last row considers the activation value of the second-to-last hidden layer without clustering (similar to~\cite{gaussianMonitor}).}
	\label{tab:overall}
	\centering
	\resizebox{0.95\textwidth}{!}{
		\begin{tabular}{l|r|r|r|r|r|r|r|r|r|r|r}
			Monitors & \rot{Wrong ID} & \rot{GTSRB} & \rot{DTD} & \rot{Gaussian} & \rot{SaltAndPepper} & \rot{Contrast} & \rot{GaussianBlur} & \rot{Invert} & \rot{Rotate} & \rot{Light} & \rot{Cifar100} \\
			\toprule
			Univariate Gaussian, 3 Clusters & 21.71 & 39.00 & 15.00 & 19.49 & 26.20 & 32.42 & 17.02 & 21.76 & 20.74 & 24.58 & 20.00 \\
			Univariate Gaussian, No Clusters & 5.20 & 19.00 & 4.00 & 3.37 & 1.56 & 11.99 & 2.53 & 3.99 & 3.18 & 9.13 & 3.00 \\
			Univariate Gaussian, \cite{gaussianMonitor} & 10.82 & 19.00 & 4.00 & 7.75 & 5.58 & 19.95 & 4.97 & 8.13 & 6.94 & 15.76 & 8.00 \\
		\end{tabular}
	}
	\resizebox{0.95\textwidth}{!}{
		\begin{tabular}{l|r|r|r|r|r|r|r|r|r|r|r}
			Monitors & \rot{Wrong ID} & \rot{CIFAR10} & \rot{DTD} & \rot{Gaussian} & \rot{SaltAndPepper} & \rot{Contrast} & \rot{GaussianBlur} & \rot{Rotate} & \rot{Invert} & \rot{Light} & \rot{CTS} \\
			Univariate Gaussian, 3 Clusters & 49.93 & 75.00 & 76.56 & 44.45 & 80.42 & 97.26 & 38.68 & 48.28 & 87.06 & 65.47 & 62.83 \\
			Univariate Gaussian, No Clusters & 21.65 & 53.91 & 64.84 & 27.90 & 46.12 & 87.87 & 20.82 & 26.42 & 74.62 & 46.28 & 36.44 \\
			Univariate Gaussian, \cite{gaussianMonitor} & 41.13 & 70.31 & 73.44 & 42.07 & 80.20 & 97.26 & 28.72 & 46.50 & 75.02 & 50.69 & 49.14 \\
		\end{tabular}
	}
\end{table}

%% file: conclusion.tex
\section{Conclusion and Future Work}

In this work, we applied the Gaussian monitor \cite{gaussianMonitor} for \ac{ood} detection for \acp{nn} to the setting of \ac{oms} detection and extended it.
Our new monitors consider the correlations of neurons by 
pre-clustering the activation values similar to~\cite{Henzinger-2020-AbstractionBasedMonitor}. 
We found that clustering the activation values before applying the Gaussian monitor outperforms the approach without clustering.
We also saw that the number of monitored neurons can be reduced while maintaining comparable detection capabilities. 
This reduces the amount of computation required in each monitoring step, which is essential for executing the monitors fast.
In future work, we want to evaluate on more datasets and investigate the possibility of combining monitoring of several layers.

%% file: appendix.tex
\section*{Appendix}
\renewcommand{\thesubsection}{\Alph{subsection}}

\subsection{Datasets} \label{app:datasets}
This section gives examples of the different datasets used for our evaluation. 
We use two different datasets as \ac{id} data, \cifarten \cite{cifar10} and GTSRB \cite{gtsrb}.

\begin{figure}
	\centering
	\input{figures/cifar10-description-images}
	\caption{Illustration of the \cifarten \cite{cifar10} dataset.}
	\label{fig:cifar10}
\end{figure}
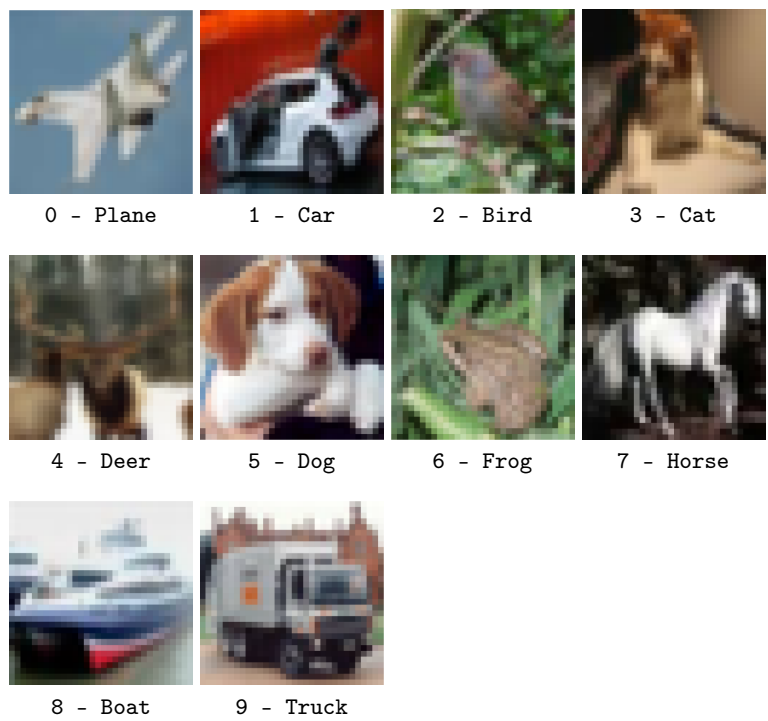

The \cifarten dataset \cite{cifar10} comprises $32 \times 32$ color images (three channels) distinguishing ten different classes. An illustration of the dataset can be found in \cref{fig:cifar10}.

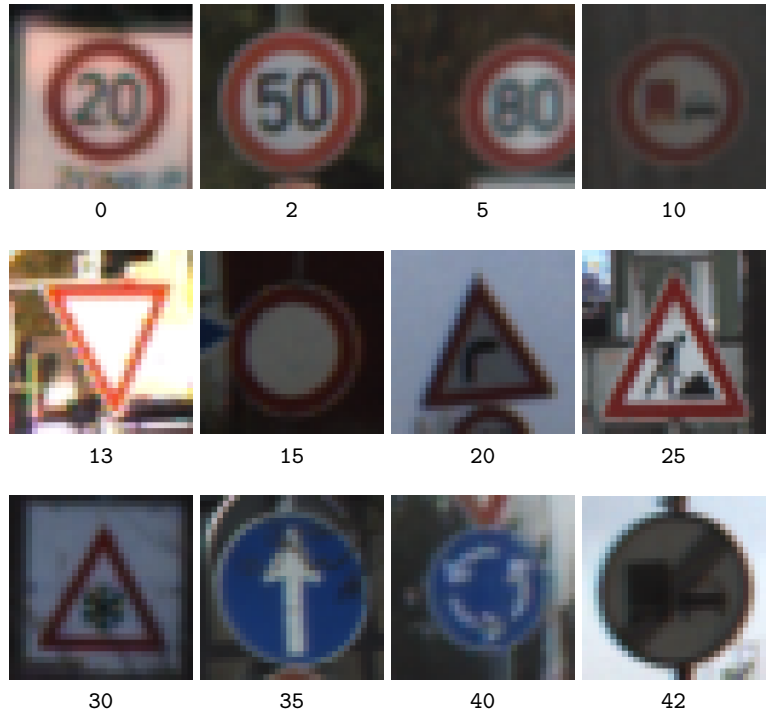
\begin{figure}
	\centering
	\input{figures/gtsrb-description-images}
	\caption{Illustration of the \gtsrb \cite{gtsrb} dataset.}
	\label{fig:gtsrb}
\end{figure}

The \gtsrb \cite{gtsrb} dataset comprises $32 \times 32$ color images (three channels), distinguishing 43 different classes of German traffic signs. An illustration of the dataset can be found in \cref{fig:gtsrb}.

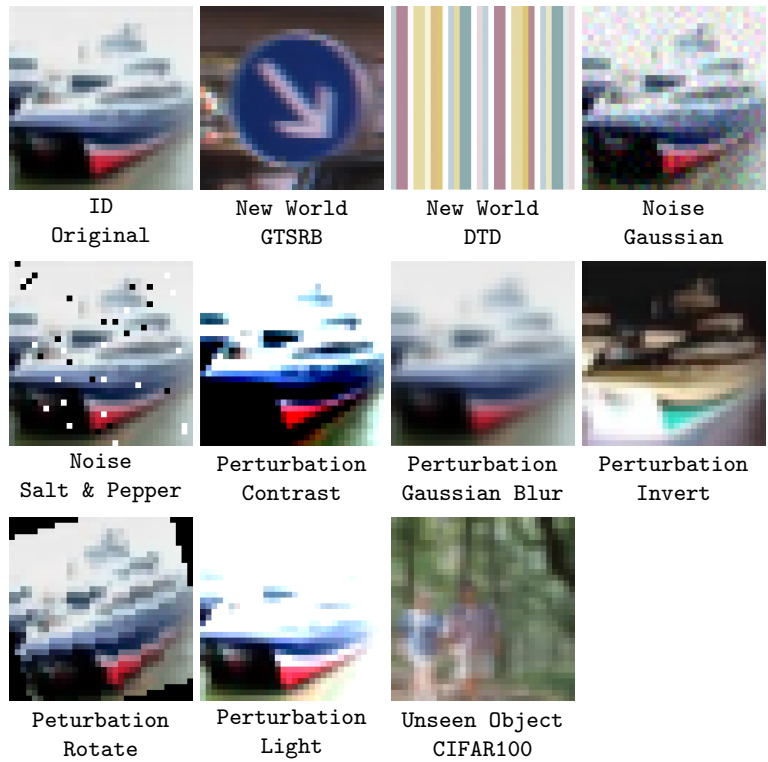
\begin{figure}
	\centering
	\input{figures/cifar10-manipulated-images}
	\caption{Illustration of various manipulations performed on the original \ac{id} data of the \cifarten dataset to create the \ac{oms} datasets.}
	\label{fig:ood_data_cifar10}
\end{figure}

\Cref{fig:ood_data_cifar10} depicts images from the \ac{id} dataset and the constructed \ac{oms} datasets for \cifarten. 
First, three datasets encode novelty: 'New World/GTSRB', 'New World/DTD', and 'Unseen Object/CIFAR100'. 
The DTD~\cite{dtd} dataset comprises images of 47 different types of textures, and CIFAR100 \cite{cifar10} is a disjoint dataset from \cifarten consisting of 100 distinct classes. 
CIFAR100 is not entirely a \textit{New World}, unlike \gtsrb and DTD, as there is an overlap in the superclasses. 
Like CIFAR10, CIFAR100 also includes images of vehicles and animals, among other types. 
Secondly, seven of them encode a covariate shift: 'Gaussian Noise', 'Salt \& Pepper Noise', 'Contrast Perturbation', 'Gaussian Blur Perturbation', 'Invert Perturbation', 'Rotate Perturbation' and 'Light Perturbation'.

We perform similar operations for the \gtsrb dataset. 
Three datasets encode novelty: 'New World/GTSRB', 'New World/

', and 'Unseen Object/CTS'.
The CTS \cite{cts} dataset includes images of various Chinese traffic signs. 
CTS is not entirely a 'New World', unlike CIFAR10 and DTD, as there is a semantical overlap. 
Like GTSRB, CTS also includes images of 'speed signs' or 'no parking signs' among other similar types. 
Secondly, seven datasets encode a covariate shift: 'Gaussian Noise', 'Salt \& Pepper Noise', 'Contrast Perturbation', 'Gaussian Blur Perturbation', 'Invert Perturbation', 'Rotate Perturbation' and 'Light Perturbation'.

\subsection{Network Architectures} \label{app:networks}
This section presents the \ac{nn} architectures used for evaluation in more detail.
The architecture for the \ac{nn} used as a classification \ac{nn} for \cifarten is available in \cref{fig:nn-architecture_cifar10}.
\Cref{fig:nn-architecture_gtsrb} shows the architecture of the \ac{nn} used for \gtsrb.
\begin{figure}
	\centering
	\begin{tabular}{|c|c|c|c|}
		\hline
		Layer index $l$& Pre-Activations $Z^l$ & Activations $A^l$ & Number of Neurons $|L^l|$\\
		\hline
		$2$ & \texttt{Conv2d()} & \texttt{ReLU()} & - \\
		$3$ & \texttt{Conv2d()} & \texttt{ReLU()} & -\\
		$4$ &  - & \texttt{MaxPool2d()} & -\\
		$5$ &  \texttt{Conv2d()} & \texttt{ReLU()} & -\\
		$6$ & - & \texttt{Dropout2d()} & -\\
		$7$ & \texttt{Conv2d()} & \texttt{ReLU()} & -\\
		$8$ &  & \texttt{MaxPool2d()} & -\\
		$9$ & \texttt{Conv2d()} & \texttt{ReLU()} & -\\
		$10$ & - & \texttt{Dropout2d()}  & -\\
		$11$ & \texttt{Linear()} & \texttt{ReLU()} & \texttt{256}\\
		$12$ & \texttt{Linear()} & \texttt{ReLU()} & \texttt{128}\\
		$13$ & \texttt{Linear()}  & \texttt{ReLU()} & \texttt{64}\\
		$14$ & \texttt{Linear()} & - & \texttt{10}\\
		\hline
	\end{tabular}
	\caption{\ac{nn} architecture for classifying images of the \cifarten dataset developed by \cite{Vishal-Ramesh-NN-Architecture-Cifar10}.}
	\label{fig:nn-architecture_cifar10}
\end{figure}

\begin{figure}
	\centering
	\begin{tabular}{|c|c|c|c|}
		\hline
		Layer index $l$& Pre-Activations $Z^l$ & Activations $A^l$ & Number of Neurons $|L^l|$\\
		\hline
		$2$ & \texttt{Conv2d()} & \texttt{ReLU(BatchNorm2d())} & - \\
		$4$ &  - & \texttt{MaxPool2d()} & -\\
		$4$ & \texttt{Conv2d()} & \texttt{ReLU(BatchNorm2d())} & -\\
		$5$ &  - & \texttt{MaxPool2d()} & -\\
		$6$ & \texttt{Linear()} & \texttt{ReLU()} & \texttt{240}\\
		$7$ & \texttt{Linear()}  & \texttt{ReLU()} & \texttt{84}\\
		$8$ & \texttt{Linear()} & - & \texttt{43}\\
		\hline
	\end{tabular}
	\caption{\ac{nn} architecture for classifying images of the \gtsrb dataset, as used by Chen et al.~\cite{Chen-2018-Semantic-Segmentation} and Hashemi et al.~\cite{gaussianMonitor}.}
	\label{fig:nn-architecture_gtsrb}
\end{figure}

\subsection{Monitoring Activation and Pre-Activation Values} \label{app:pre-act}
We evaluate whether to monitor activation or pre-activation values.
We focus on hidden layers close to the output as the literature, e.g.,~\cite{gaussianMonitor,Corbi-2019-QualityScores,Henzinger-2020-AbstractionBasedMonitor}, has found them to be the most useful ones for monitoring.
\Cref{tab:univariate_gaussian_best_on_different_layers} depicts the \ac{tpr} of the best monitor configuration for each pre-activation and activation values $Z^{12}, A^{12}, Z^{13} \text{ and } A^{13}$ for the clustered Gaussian monitor on \cifarten.
We optimized the number of clusters 
beforehand.
The results indicate that monitoring the activation values $Z^{12}$ and $Z^{13}$ is more effective than monitoring the activation values  $A^{12}$ and $A^{13}$.
Furthermore, monitoring $Z^{13}$ instead of $Z^{12}$ improves performance in all test scenarios for the clustered Gaussian monitor.
\input{plots/cifar10/un-g-all-layers-plot.tex}

%% file: figures/cifar10-description-images.tex
    \begin{tabular}{cccc}
        \includegraphics[width=0.20\linewidth]{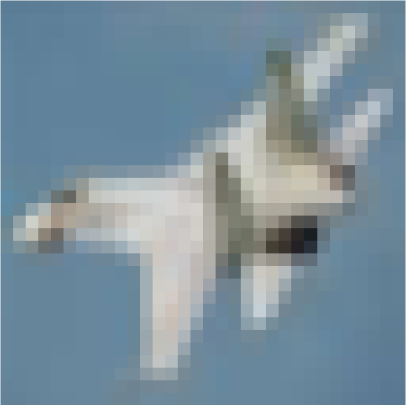} & \includegraphics[width=0.20\linewidth]{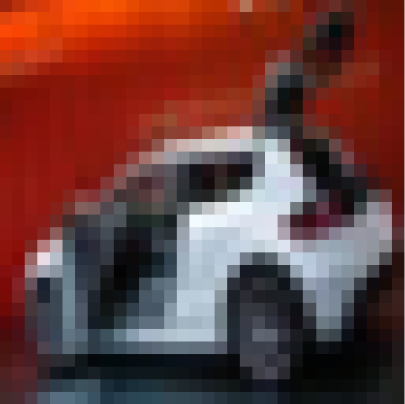} & \includegraphics[width=0.20\linewidth]{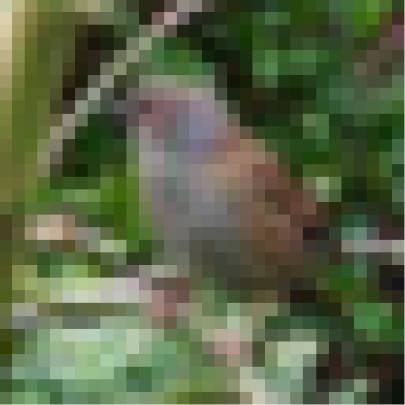} &
        \includegraphics[width=0.20\linewidth]{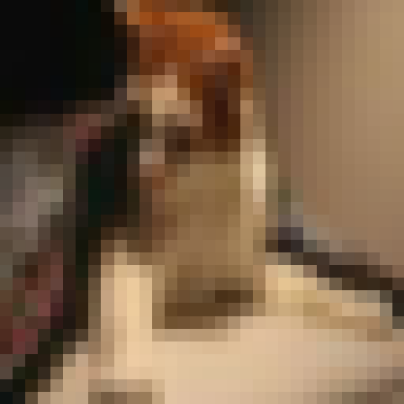}\\
        \texttt{0 - Plane} & \texttt{1 - Car}  & \texttt{2 - Bird}  & \texttt{3 - Cat}\\[1em]
        \includegraphics[width=0.20\linewidth]{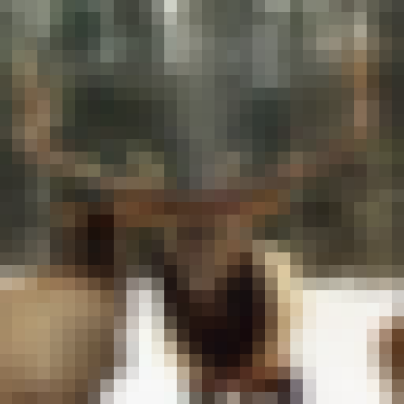} & \includegraphics[width=0.20\linewidth]{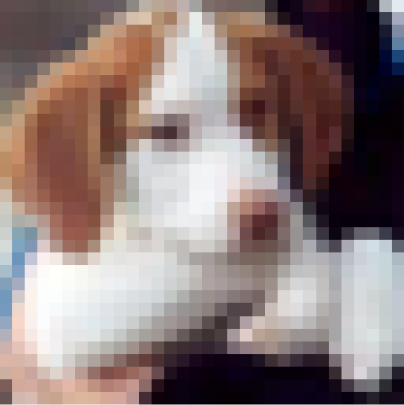} & \includegraphics[width=0.20\linewidth]{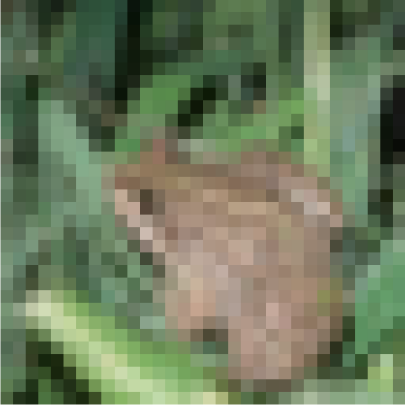} &
        \includegraphics[width=0.20\linewidth]{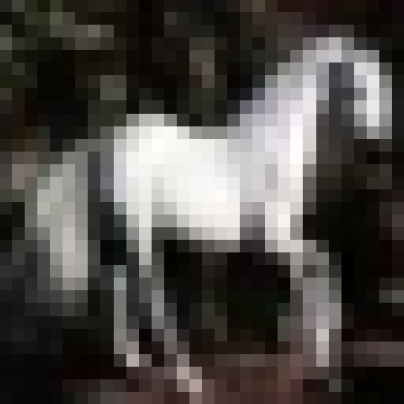} \\
        \texttt{4 - Deer}  & \texttt{5 - Dog} &\texttt{6 - Frog} & \texttt{7 - Horse} \\[1em]
        \includegraphics[width=0.20\linewidth]{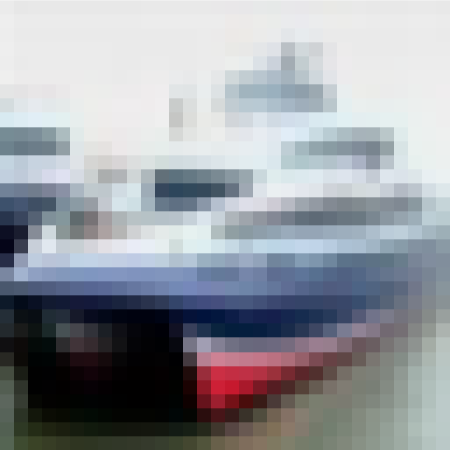}&
        \includegraphics[width=0.20\linewidth]{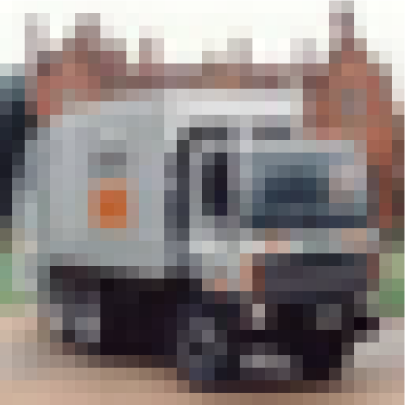} &\\
           \texttt{8 - Boat} & \texttt{9 - Truck}   \\[1em]
    \end{tabular}

%% file: figures/gtsrb-description-images.tex
\begin{tabular}{cccc}
        \includegraphics[width=0.20\linewidth]{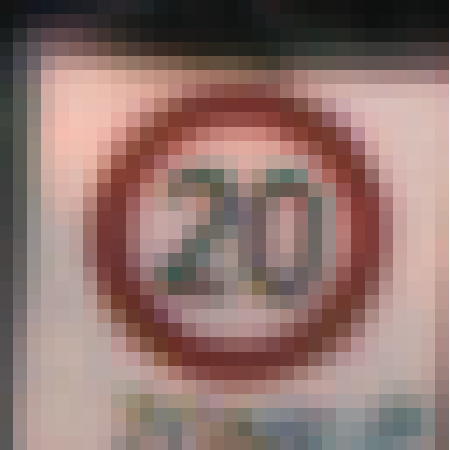} & 
        \includegraphics[width=0.20\linewidth]{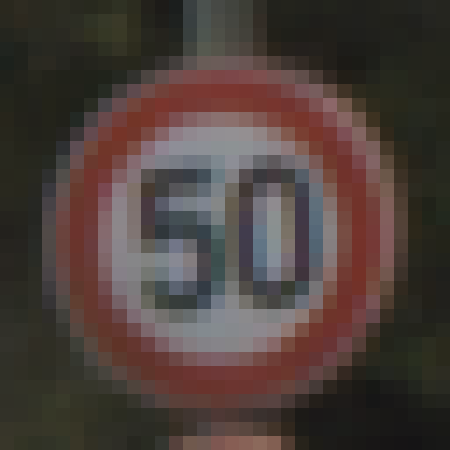} & 
\includegraphics[width=0.20\linewidth]{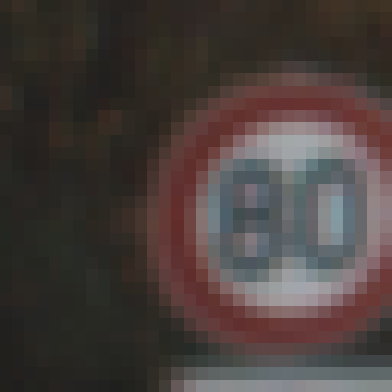} & \includegraphics[width=0.20\linewidth]{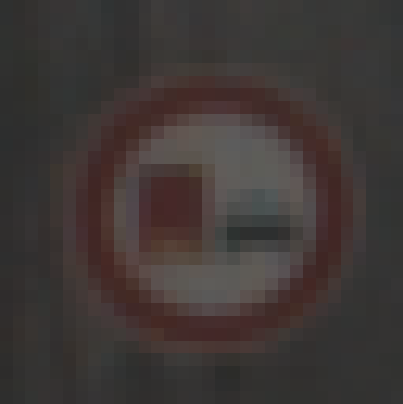} \\
        \texttt{0} & \texttt{2}  & \texttt{5}  & \texttt{10}\\[1em]
                \includegraphics[width=0.20\linewidth]{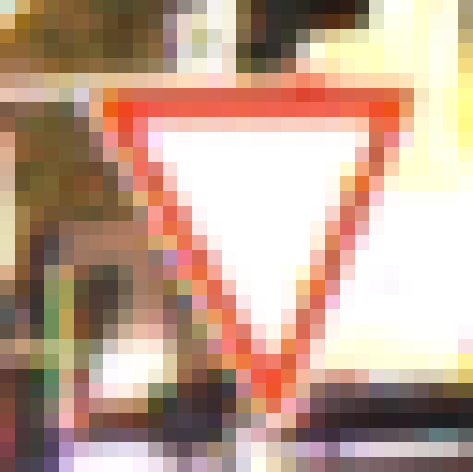} & 
                   \includegraphics[width=0.20\linewidth]{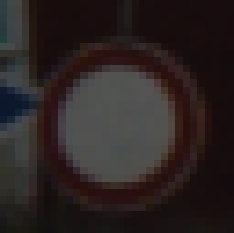}& 
        \includegraphics[width=0.20\linewidth]{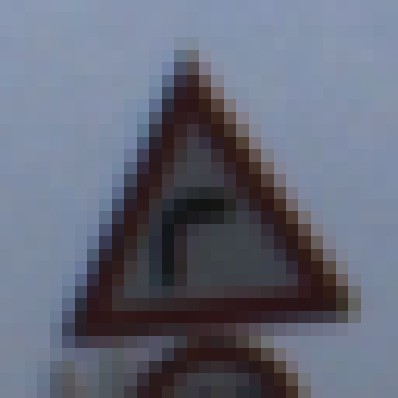} & \includegraphics[width=0.20\linewidth]{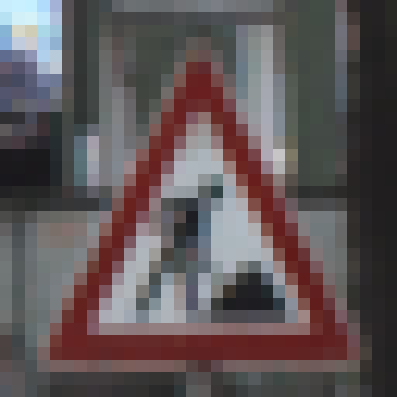}  \\
        \texttt{13}  & \texttt{15} &\texttt{20} & \texttt{25} \\[1em]
        \includegraphics[width=0.20\linewidth]{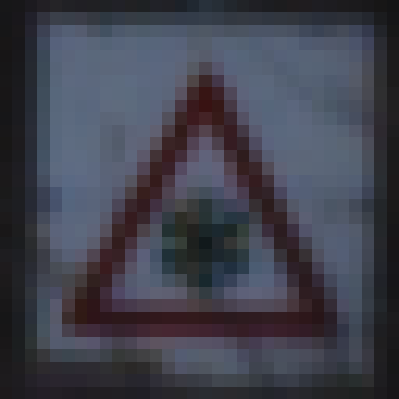} &
        \includegraphics[width=0.20\linewidth]{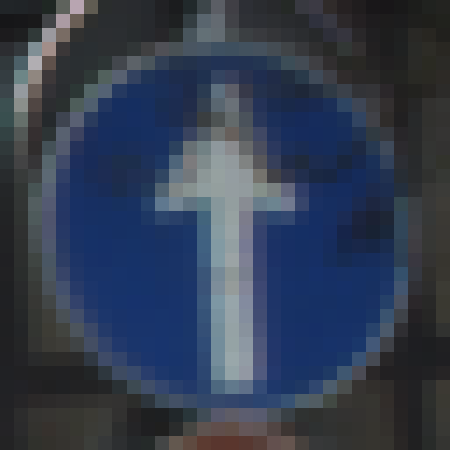} &
        \includegraphics[width=0.20\linewidth]{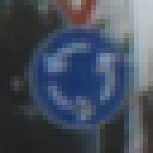}&
        \includegraphics[width=0.20\linewidth]{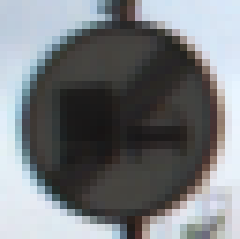}\\
          \texttt{30} & \texttt{35} & \texttt{40} & \texttt{42}   \\
\end{tabular}

%% file: figures/cifar10-manipulated-images.tex
    \begin{tabular}{cccc}
       \includegraphics[width=0.20\linewidth]{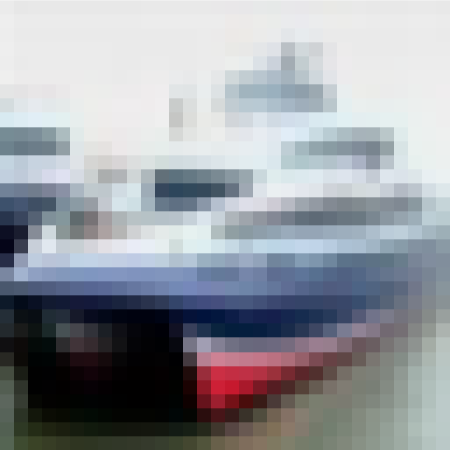} & 
    \includegraphics[width=0.20\linewidth]{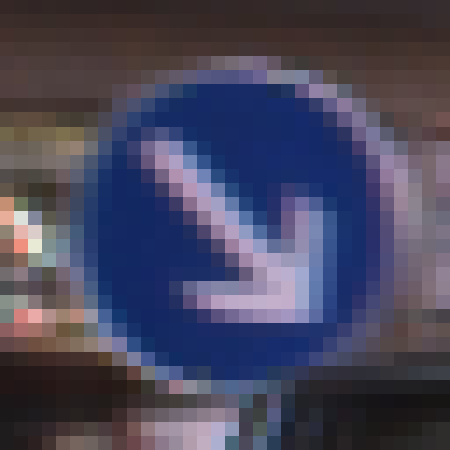} & 
    \includegraphics[width=0.20\linewidth]{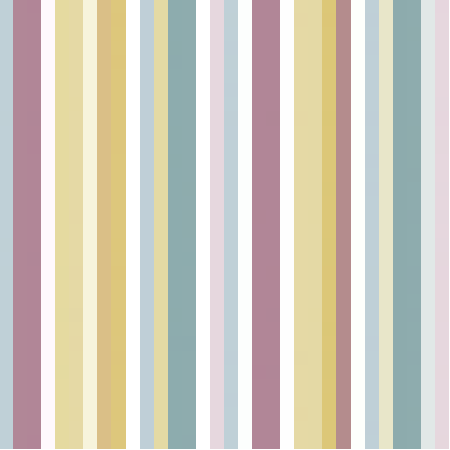} & \includegraphics[width=0.20\linewidth]{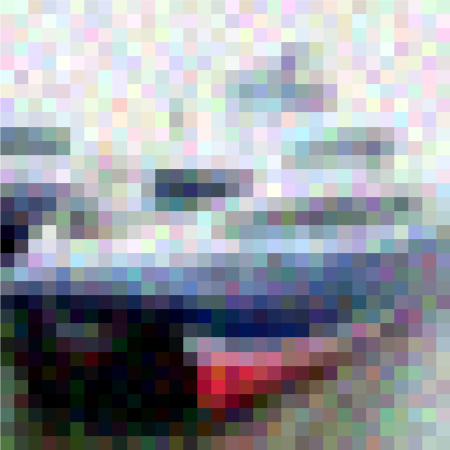} \\
     \parbox{0.20\linewidth}{\centering\texttt{ID \\ Original}} & 
     \parbox{0.20\linewidth}{\centering\texttt{New World \\ GTSRB}} &  
     \parbox{0.20\linewidth}{\centering\texttt{New World \\ DTD}} &
     \parbox{0.20\linewidth}{\centering\texttt{Noise \\  Gaussian}}\\ [1em]
    
    \includegraphics[width=0.20\linewidth]{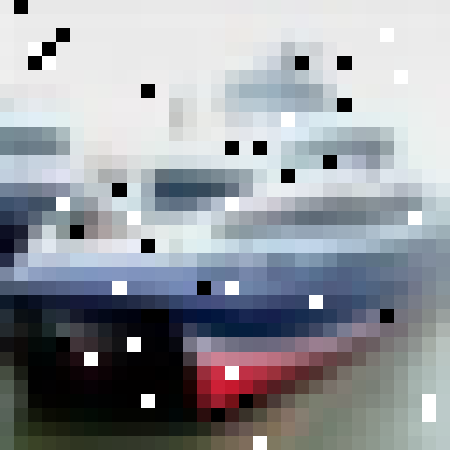} & 
    \includegraphics[width=0.20\linewidth]{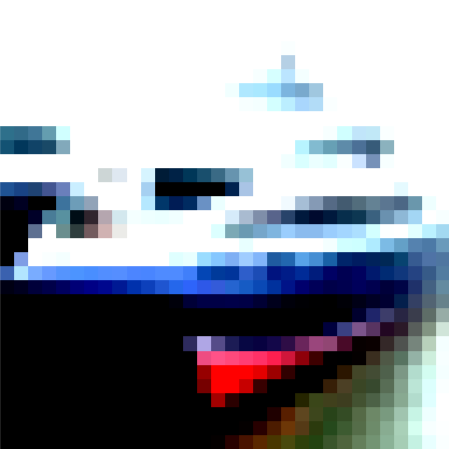} &
     \includegraphics[width=0.20\linewidth]{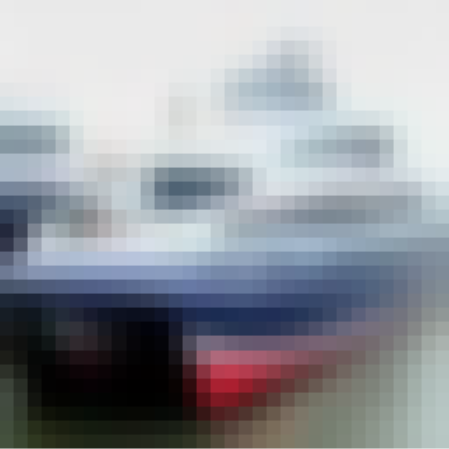} & 
    \includegraphics[width=0.20\linewidth]{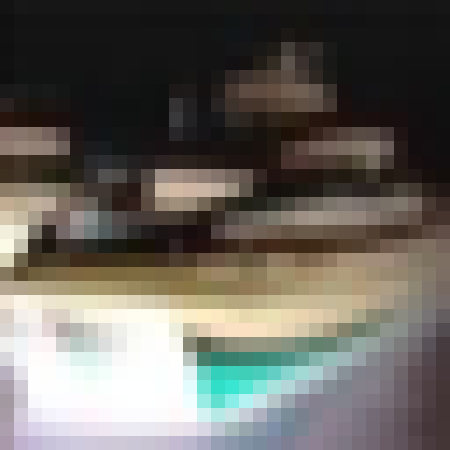}  \\
    \parbox{0.20\linewidth}{\centering\texttt{Noise \\ Salt \& Pepper}} & 
     \parbox{0.20\linewidth}{\centering\texttt{Perturbation \\ Contrast}} &  
     \parbox{0.20\linewidth}{\centering\texttt{Perturbation \\ Gaussian Blur }} &
     \parbox{0.20\linewidth}{\centering\texttt{Perturbation \\ Invert}}\\ [1em]

    \includegraphics[width=0.20\linewidth]{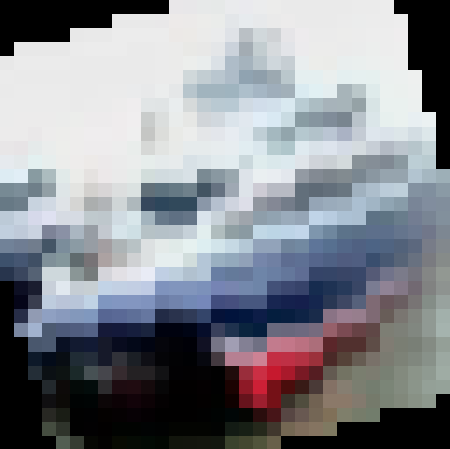} &
     \includegraphics[width=0.20\linewidth]{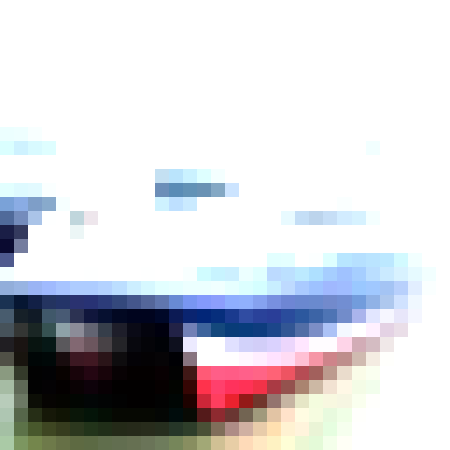} & 
    \includegraphics[width=0.20\linewidth]{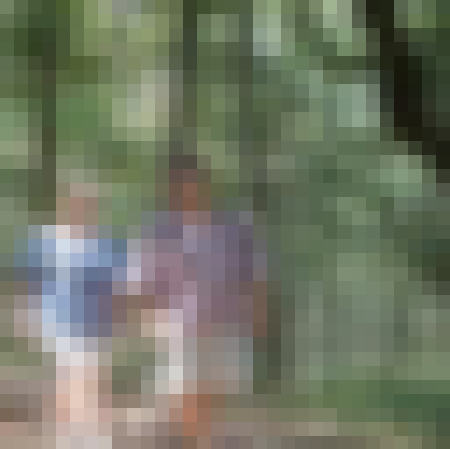} & \\
     \parbox{0.20\linewidth}{\centering\texttt{Peturbation \\ Rotate}} & 
     \parbox{0.20\linewidth}{\centering\texttt{Perturbation \\ Light}} &  
     \parbox{0.20\linewidth}{\centering\texttt{Unseen Object \\ CIFAR100}} & \\[1em]
  
    \end{tabular}

%% file: plots/cifar10/un-g-all-layers-plot.tex
\begin{table}
	\caption{Comparison of the \ac{tpr} for the best configuration of the clustered Gaussian monitor trained on the pre-activation ($Z^{12}, Z^{13}$) and activation values ($A^{12}, A^{13}$) of the last two hidden layers for the \cifarten \ac{nn}.}
	\label{tab:univariate_gaussian_best_on_different_layers}
	\centering
	\resizebox{0.9\textwidth}{!}{
		\begin{tabular}{l|r|r|r|r|r|r|r|r|r|r|r}
			Monitors & \rot{Wrong ID} & \rot{GTSRB} & \rot{DTD} & \rot{Gaussian} & \rot{SaltAndPepper} & \rot{Contrast} & \rot{GaussianBlur} & \rot{Invert} & \rot{Rotate} & \rot{Light} & \rot{Cifar100} \\
			\toprule
			$Z^{12}$, 3 \ Clusters\; & 11.78 & 37.00 & 7.00 & 8.75 & 10.69 & 26.77 & 5.18 & 9.57 & 8.27 & 17.51 & 12.00 \\
			$A^{12}$, 2 \ Clusters & 3.27 & 7.00 & 1.00 & 2.28 & 1.05 & 7.59 & 0.96 & 2.02 & 1.55 & 4.98 & 1.00 \\
			$Z^{13}$, 3 \ Clusters & 21.71 & 39.00 & 15.00 & 19.49 & 26.20 & 32.42 & 17.02 & 21.76 & 20.74 & 24.58 & 20.00 \\
			$A^{13}$, 2 \ Clusters & 1.58 & 3.00 & 0.00 & 1.28 & 0.44 & 4.04 & 0.33 & 0.92 & 0.64 & 2.45 & 1.00 \\
		\end{tabular}
	}
\end{table}